\documentclass[sigconf]{aamas} 

\usepackage{balance} 

\setcopyright{ifaamas}
\acmConference[AAMAS '23]{}
\copyrightyear{}
\acmYear{}
\acmDOI{}
\acmPrice{}
\acmISBN{}

\title[Reward Relabelling]{STIR$^2$: Reward Relabelling for combined Reinforcement and Imitation Learning on sparse-reward tasks}

\author{Jesus Bujalance}
\affiliation{
  \institution{MINES ParisTech, PSL University, Center for robotics}
  \city{Paris}
  \country{France}}
\email{jesus.bujalance\_martin@mines-paristech.fr}

\author{Fabien Moutarde}
\affiliation{
  \institution{MINES ParisTech, PSL University, Center for robotics}
  \city{Paris}
  \country{France}}

\begin{abstract}
In the search for more sample-efficient reinforcement-learning (RL) algorithms, a promising direction is to leverage as much external off-policy data as possible. For instance, expert demonstrations. In the past, multiple ideas have been proposed to make good use of the demonstrations added to the replay buffer, such as pretraining on demonstrations only or minimizing additional cost functions.
We present a new method, able to leverage both demonstrations and episodes collected online in any sparse-reward environment with any off-policy algorithm. Our method is based on a reward bonus given to demonstrations and successful episodes (via relabeling), encouraging expert imitation and self-imitation.
Our experiments focus on several robotic-manipulation tasks across two different simulation environments. We show that our method based on reward relabeling improves the performance of the base algorithm (SAC and DDPG) on these tasks. 
Finally, our best algorithm STIR$^2$ (Self and Teacher Imitation by Reward Relabeling), which integrates into our method multiple improvements from previous works, 
is more data-efficient than all baselines. 
\end{abstract}

\newcommand{\BibTeX}{\rm B\kern-.05em{\sc i\kern-.025em b}\kern-.08em\TeX}
\usepackage{amsmath}
\usepackage{graphicx}
\usepackage{subcaption}
\usepackage{dsfont}
\usepackage[ruled,vlined]{algorithm2e}

\begin{document}


\pagestyle{fancy}
\fancyhead{}


\maketitle 


\section{Introduction}
\label{introduction}


Despite having known great success, RL algorithms still need to become more sample-efficient, particularly for robotics where it is much harder to deploy an algorithm outside of simulation.
In this work we focus on off-policy RL, and present a way to leverage offline data in the form of 
expert 
demonstrations. 
Our method is based on the observation that, in hindsight, a successful episode of collected experience is in fact a demonstration, so it should receive the same treatment. In particular, we propose to add a reward bonus to transitions coming from both demonstrations and successful episodes. Our approach provides a simple way of tying positive rewards and desired behaviour, without any task-specific reward shaping.

We focus on multiple manipulation tasks with sparse rewards, such as reaching a target, pushing a button, and sliding a block to a target position. 
We instantiate our approach with Soft Actor-Critic (SAC) \cite{haarnoja2018soft} and Deep Deterministic Policy Gradient (DDPG) \cite{DBLP:journals/corr/LillicrapHPHETS15}, and compare it to three other algorithms \cite{DBLP:journals/corr/VecerikHSWPPHRL17, nair2018overcoming, DBLP:journals/corr/abs-2012-11989}, presented in detail in Section \ref{related_work}. Just like ours, these methods are task-agnostic and generic, in the sense that they can be applied to any continuous-action off-policy algorithm with some minor modifications.
The contributions of this paper are the following:
\begin{itemize}
    \item We introduce a new method that consists in giving a reward bonus to demonstrations and relabeling successful episodes as demonstrations. We test it across different tasks and algorithms, and show that it greatly improves upon the base algorithm. 
    \item Similarly to previous works such as Rainbow \cite{hessel2018rainbow}, we propose a new algorithm STIR$^2$ that integrates into our method multiple improvements from other works, and show that it outperforms all baselines.
\end{itemize}





\section{Related work}
\label{related_work}


\bigbreak
\textbf{Learning for robotics.}
Successful algorithms in robotic manipulation tend to take advantage of the particular structure of the tasks they wish to solve. The recent RAPS \cite{dalal2021accelerating} alters the action space by manually specifying a library of parameterized robot action primitives, and the recent C2F-ARM \cite{james2021coarse} relies on point cloud inputs and a clever pre-processing pipeline to achieve impressive results from just a handful of demonstrations. 
Outside of RL, Transporters Networks \cite{zeng2020transporter} exploit spatial symmetries and RGB-D sensors to solve a variety of tasks from just a few samples, and NTP \cite{DBLP:journals/corr/abs-1710-01813} and the more recent NTG \cite{huang2019neural} use clever representations to execute a hierarchy of movement primitives that solve complex sequential tasks from demonstrations. 
Many successful algorithms also come from data-driven fields such as offline RL (e.g. MT-Opt \cite{kalashnikov2021mt}) and model-based RL, like \cite{finn2016deep} or \cite{zhang2019solar} for models in a learnt latent space, and \cite{finn2017deep} or \cite{schmeckpeper2020learning} for models directly in image space.

\bigbreak
\textbf{Learning from demonstrations.}
The most straight-forward variant of Imitation Learning (IL) is behaviour cloning (BC), where we directly look for a policy that acts like the expert by solving a supervised learning problem, such as in \cite{bojarski2016end}. 
More robust algorithms have been proposed, like Dagger \cite{ross2011reduction}, or the recent Implicit BC \cite{florence2021implicit}, achieving impressive results with energy-based models.
Another variant of IL is inverse RL, where we try to infer the reward function that the expert was most likely trying to maximize, while optionally jointly learning a policy. 
The most recent examples are based on adversarial optimization, such as GCL \cite{finn2016guided}, which is similar to GAIL \cite{ho2016generative}.

\bigbreak
\textbf{Self-Imitation Learning.}
Having an expert to learn from can be very helpful, but such a luxury is not always available. Instead, we can apply IL methods to the experiences collected by the agent in the environment, to further supervise its behaviour.
\cite{Oh2018SIL} introduced Self-Imitation Learning (SIL), where an additional loss function pushed the agent to imitate its own decisions in the past only when they resulted in larger returns than expected. 
Further works such as Self-Imitation Advantage Learning (SAIL) \cite{DBLP:journals/corr/abs-2012-11989} followed. We will refer to as SAC-SAIL and DDPG-SAIL to the implementations built upon SAC and DDPG respectively.

\bigbreak
\textbf{Reward bonus.}
One major challenge in RL is exploration, and one common approach to encourage it is to augment the environment reward with an additional bonus. Many previous works rely on some sort of optimistic exploration: assume the unknown to be good. A simple idea is counting state occurrences, even if the space is too large or continuous via pseudo-counts \cite{bellemare2016unifying} or hashing \cite{tang2017exploration}.
Another idea is to use any type of prediction error, which can quantify the novelty of new experience. For instance, in \cite{pathak2017curiosity} they learn a forward model to predict future states, and in RND \cite{burda2018exploration} they predict the output of a fixed randomly initialized neural network on the current observation.
More generally, reward bonuses are used to provide further supervision to the agent. For instance,
\cite{shelhamer2016loss} presents a variety of self-supervised losses such as simple proxies of the reward (e.g. predict the sign), the dynamics (or inverse dynamics), or based on reconstruction errors (e.g. auto-encoders).

\bigbreak
\textbf{Relabeling past experience.}
Since off-policy RL algorithms can theoretically use data coming from any policy, a natural idea was to share data between tasks in a multi-task setting. An even better idea came in HER \cite{NIPS2017_453fadbd}, where the authors pointed out that if we accidentally solve one task when trying to perform another task, that experience is still optimal if we relabel the goal that was initially intended. Similar and more general works followed, such as GCSL \cite{ghosh2019learning}, Generalized Hindsight \cite{li2020generalized}, and HIPI \cite{eysenbach2020rewriting}, which reframes the relabeling problem as inverse RL. RCP \cite{DBLP:journals/corr/abs-1912-13465} extended the idea to the single-task setting, by learning a policy conditioned on the trajectory return.

\bigbreak
\textbf{Learning from both demonstrations and reinforcement learning.}
Demonstrations can be used to design the reward, guide exploration, augment
the training data, initialize policies, etc. In NAC \cite{gao2018reinforcement}, the demonstrations are used as the only training data during the first iterations. In \cite{zhu2018reinforcement} the demonstrations are used for an additional imitation-based reward. In DAPG \cite{DBLP:journals/corr/abs-1709-10087} they are used to augment the policy gradient equation. 
In DAC \cite{liu2020demonstration}, they introduce a novel objective based on an augmented reward, the larger the closer the policy to the expert policy.
We will focus on three algorithms that can be applied to any continuous-action off-policy algorithm with minor modifications.

DDPG-fD and SAC-fD \cite{DBLP:journals/corr/VecerikHSWPPHRL17} (originally DDPGfD based on DDPG) introduces three main ideas: transitions from demonstrations are added to the replay buffer, demonstration data is sampled more often from the buffer, and a mix of 1-step and n-step return losses is used.

DDPG-BC and SAC-BC \cite{nair2018overcoming} (has no name and originally based on DDPG) also introduces three main ideas: transitions from demonstrations are added to a separate additional replay buffer, an auxiliary behaviour cloning loss is applied to samples from this buffer,
and some episodes are reset to a state sampled uniformly from a demonstration.
The authors additionally present other ideas regarding the multi-goal setting which we will not cover in this work.

SQIL \cite{reddy2020sqil} is actually a pure IL algorithm, since the reward signal is supposed unknown, but we present it here since it also incorporates demonstration data into the replay buffer. The replay buffer is initially filled with demonstrations where the rewards are always $r = 1$, and new experiences collected by the agent are added with reward $r = 0$. 


\section{Background}
\label{background}

\textbf{Reinforcement Learning.}
RL provides a framework where an agent interacts with an environment by performing an action and observing a feedback signal (reward $r$) and the new state of the environment. 
The goal is to find the policy $\pi$ (function mapping states $s$ to actions $a$) that maximizes the discounted cumulative reward (the parameter $\gamma\leq 1$ makes future rewards smaller). Two popular algorithms for continuous-action spaces are SAC \cite{haarnoja2018soft} and DDPG \cite{DBLP:journals/corr/LillicrapHPHETS15}.
\begin{equation}
\notag
    \pi^* = \arg\max\limits_\pi \mathbb{E}_{\tau \sim p_\pi(\tau)} \left[ \sum\limits_{k=0}^{T} \gamma^k r(s_{1+k},a_{1+k}) \right] 
    \label{objective}
\end{equation}
The \textit{Q-function} $Q^\pi(s_t,a_t) = \sum\limits_{t'=t}^{T} \mathbb{E}_{p_\pi} \left[ \gamma^{t'-t} r(s_{t'},a_{t'})|s_t,a_t \right]$ is defined as the reward-to-go from the state $s_t$ if we pick the action $a_t$ and then follow $\pi$. This function obeys the Bellman equation:
\begin{equation}
\notag
    Q^\pi(s,a) = r + \gamma\mathbb{E}_{s' \sim p(\cdot|s,a),a' \sim \pi(\cdot|s')} \left[ Q^\pi(s',a')\right]
\end{equation}

\bigbreak
\textbf{Deep Deterministic Policy Gradient.} The goal is to learn a deterministic policy $\mu_\theta$.
To train the critic, we can approximate the right-hand expectation of the Bellman equation with samples, set it equal to $y$, and minimize the MSBE loss on a parameterized $Q_\phi$ (in practice, two additional \textit{target} approximators $Q_{\phi'}$ and $\mu_{\theta'}$ are used to compute the targets $y$):
\begin{equation}
\notag
    \mathcal{L}(Q_\phi) = \underset{(s,a,r,s',d) \sim {\mathcal D}}{{\mathbb E}}\left[
    ( Q_{\phi}(s,a) - y )^2
    \right]
\label{ddpg}
\end{equation}
To train the actor, simple gradient descent w.r.t. $\theta$ on the following loss:
\begin{equation}
\notag
    \mathcal{L}(\mu_\theta) = -\underset{s \sim {\mathcal D}}{{\mathbb E}}\left[
    Q_{\phi}(s,\mu_\theta(s))
    \right]
\label{ddpg-actor}
\end{equation}

\bigbreak
\textbf{Soft Actor-Critic.}
In classic RL, the optimal policy is always deterministic under full observability, but stochastic policies have interesting properties: better exploration and robustness (due to wider coverage of states), and multi-modality. This new objective promotes stochasticity by maximizing the entropy $\mathcal{H}$ of the policy ($\alpha$ is a hyper-parameter, and we omit $\gamma$ for simplicity):
\begin{equation}
\notag
    \pi^* =  \arg\max_\pi \sum\limits_{t=1}^{T} \mathbb{E}_{(s_t,a_t) \sim p_\pi} \left[ r(s_t,a_t) + \alpha \mathcal{H}(\pi(\cdot|s_t))\right] 
\end{equation}
A new Q-function (slightly different) is derived and follows the \textit{soft} Bellman equation:
\begin{equation}
\notag
    Q^\pi(s,a) = r + \gamma\mathbb{E}_{s' \sim p(\cdot|s,a),a' \sim \pi(\cdot|s')} \left[ Q^\pi(s',a') - \alpha\log\pi(a'|s')\right]
\end{equation}
Similar to DDPG, to train the critic we can approximate the right-hand expectation with samples, set it equal to $y$, and minimize the MSBE loss on a parameterized $Q_\phi$ (in practice, two approximators $Q_{\phi_1}$ and $Q_{\phi_2}$ are trained, with their respective \textit{target} versions $Q_{\phi'_1}$ and $Q_{\phi'_2}$).

To train the actor $\pi_\theta$, the actor loss is derived from the reparameterization trick to compute samples $\Tilde{a}_\theta(s,\epsilon) = \mu_\theta(s) + \sigma_\theta(s)\epsilon$, where $\epsilon$ is some random noise:
\begin{equation}
\notag
    \mathcal{L}(\pi_\theta) = -\underset{s \sim {\mathcal D}, \epsilon\sim{\mathcal N}}{{\mathbb E}}\left[
    \min_{j=1,2} Q_{\phi_j}(s,\Tilde{a}_\theta(s,\epsilon))
    - \alpha\log\pi_\theta(\Tilde{a}_\theta(s,\epsilon)|s)
    \right]
\label{sac-actor}
\end{equation}


\section{Method}
\label{method}

We propose SAC-R$^2$ and DDPG-R$^2$, acronyms for SAC with Reward Relabeling and DDPG with Reward Relabeling respectively, a straight-forward method that could be implemented to any other off-policy RL algorithm with sparse rewards.
Let $R$ be the sparse reward from the environment, $b$ the reward bonus of our method, and $L$ the amount of transitions that will receive the bonus.
First, we add demonstration data to the buffer: the last transition of each expert trajectory is given the sparse reward $R$, and the last non-final $L$ transitions are given a reward equal to $b$.
Then, as the agent explores the environment during training, every new successful episode is relabeled the same way: the last $L$ transitions leading to the sparse reward are assigned a reward equal to $b$.
Intuitively, our method helps propagate the signal of the sparse reward by explicitly turning zero rewards into positive rewards, rather than entirely relying on bootstrapping rare future rewards.

\begin{figure}[h]
\begin{center}
\includegraphics[width=0.5\textwidth]{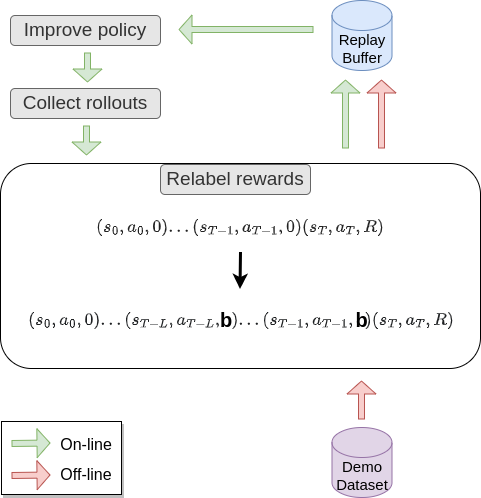}
\end{center}
\caption{Reward Relabeling procedure.}
\label{procedure}
\end{figure}

\bigbreak
\textbf{Reward bonus to demonstrations.}
The first part of our algorithm is most similar to SQIL \cite{reddy2020sqil}. 
Intuitively, it gives the agent an incentive to imitate the expert. Their paper shows theoretical connections between SQIL and regularized behaviour cloning.
One important difference is that SQIL is a pure IL algorithm, while our method learns from both the reward bonuses and the reward from the environment. Also, SQIL gives an incentive to avoid states that were not in the demonstration data, which could potentially be harmful if those states led to successful behaviour.

\bigbreak
\textbf{Reward bonus to successful episodes.}
The relabeling part of our algorithm tries to mitigate this issue and is most similar to SIL \cite{Oh2018SIL} or SAIL \cite{DBLP:journals/corr/abs-2012-11989}. In our method, the self-imitation is achieved by effectively treating successful episodes as if they were demonstrations. In order to give a reward bonus to successful episodes, we need to wait for the episodes to end first. We follow the idea presented in HER \cite{NIPS2017_453fadbd} to relabel past experiences and modify the transitions' rewards. To the best of our knowledge, HIPI \cite{eysenbach2020rewriting} is the only existing method to also modify the rewards this way. 
While their method relies on inverse RL to tackle the more general multi-task setting, ours is more straight-forward for the simpler single-task setting with sparse rewards.

\subsection{Decay and Hyper-parameters}
\label{theory}

One issue with our method so far, is that it might change the optimal policy of the problem. Intuitively, the bonus encourages a policy that requires at least $L$ steps to reach the goal, which might not take the shortest path available. We want our method to converge to the optimal policy of the sparse reward problem.

As stated in \cite{ng1999policy}, in order to still converge to the optimal policy of the original problem one can only add a reward term such as, for a given transition between two states, the term is expressible as in the difference in value of an arbitrary potential function applied to those states. Our reward bonus cannot be expressed as such since it depends on the time-step of the states. To ensure that our method converges to the optimal policy, we decide to use a decay that eventually causes the bonus to completely disappear. By doing so, our method operates in two steps: An initial RL training with reward bonuses that might not converge to the optimal policy, followed by a second RL training without any reward bonuses which should converge to the optimal policy. In practice, on top of not adding any future bonuses during data collection, we also have to remove the reward bonuses from the transitions stored in the buffer. Since the environment's rewards are fully sparse, this simply requires to ignore any reward from any non-final transition sampled from the buffer.
For our experiments, we use a simple linear decay that quickly turns the bonus to 0 when the success rate stops improving.

Our method has two hyper-parameters to tune, $b$ and $L$. $L$ represents how far the sparse reward should be propagated into the past, and is similar to other parameters in RL such as the $n$ of the n-step look-ahead in Q-learning. 
Why not relabel all time-steps? We do not wish to reward the early transitions of overly long episodes since they probably did not help to the completion of the task. Also, the cumulative rewards would greatly vary between episodes.

\begin{figure}[h]
\begin{center}
\includegraphics[width=0.3\textwidth]{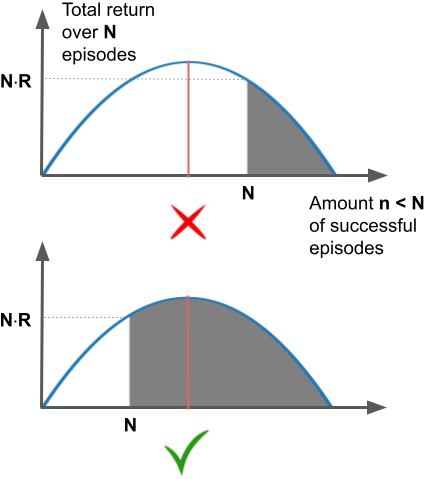}
\end{center}
\caption{Reward obtained as a function of the agent's success. The more successful the agent, the smaller the bonus, until recovering the fully sparse reward at $100\%$ success rate. We want to avoid the situation on the top, where the highest cumulative reward is obtained for a smaller success rate.}
\label{decay}
\end{figure}

How about $b$? We propose to adjust the value of $b$ during training based on the agent's recent success. This also acts as a more natural decay during the entire training process. Let's place ourselves in the episodic undiscounted scenario.
Let $\zeta$ be the fraction of successful episodes over the last 100 episodes, and $r$ the total reward obtained over a successful episode $\tau$.

\begin{equation}
    r(\tau) = R + L \cdot b \cdot (1 - \zeta)
\label{pseudo-decay}
\end{equation}

Intuitively, the more the agent struggles, the larger the bonus to help guide it.
We can study the behaviour of this reward function in the case where the agent reaches a constant final success rate. Let $f$ the cumulative reward obtained over $N$ runs, and $n$ the number of successful episodes.

\begin{equation}
\notag
    f(n) = n(R + L \cdot b \cdot (1 - \frac{n}{N})) = n(R+b \cdot L) - n^2 \frac{b \cdot L}{N}
\end{equation}


The cumulative reward $f$ is shown in Figure \ref{decay}. If not careful, the optimal behaviour for this reward can correspond to a policy that has not actually reached a success rate of $100\%$. There is a simple condition to avoid this: $N \leq \arg\max f $ if and only if $b \cdot L \leq R$. We are free to chose $b$ and $L$ as we want, as long as their product, corresponding to the total reward coming from bonuses in an episode, is smaller than the final sparse reward. A similar condition holds for the discounted case.

\begin{equation}
\label{condition}
\begin{split}
    \text{undiscounted}\quad &
    b \cdot L \leq R
    \\
    \text{discounted}\quad &
    b \left( \sum_{i=0}^{L-1}\gamma^i \right) \leq R \cdot \gamma ^L
\end{split}
\end{equation}

Regardless of whether we use this decay, it seems like an appropriate condition: $R$ should prevail over the bonuses.
This condition should also help to ensure that the policy can recover to the optimal policy once the bonuses have disappeared from the training.


\subsection{Full algorithm}
\label{full}

In this section we introduce SAC-STIR$^2$ and DDPG-STIR$^2$ (acronym for Self and Teacher Imitation by Reward Relabeling), two new algorithms that integrate four main improvements into the base SAC and DDPG algorithms. All these improvements could also be added to any other continuous-action
off-policy RL algorithm.
Let's define $\mathcal{L}(\pi_\theta)$ and $\mathcal{L}(Q_\phi)$ as the usual SAC and DDPG losses for policy and Q-function respectively.
Other than the reward-relabeling hyper-parameter $L$ discussed earlier, the full algorithm introduces four additional hyper-parameters: $n$, $\lambda_n$, $\lambda_\text{BC}$, and $\lambda_\text{SAIL}$. We now explain each improvement in more details.


\bigbreak
\textbf{Reward Relabeling.}

Our method as presented in section \ref{method}.

\bigbreak
\textbf{N-step loss.}
Following SAC-fD \cite{DBLP:journals/corr/VecerikHSWPPHRL17}, we use the n-step loss $\mathcal{L}_n$ to train the critic. This loss is a modified version of the standard Q-function loss $\mathcal{L}(Q_\phi) = \mathcal{L}_1(Q_\phi)$, with n-step returns replacing the immediate reward. Using a larger look-ahead $n$ can be particularly useful for tasks with sparse rewards, since it increases the chances of encountering a reward. For instance, for the simpler DDPG case we have: 

\begin{equation}
\notag
\mathcal{L}_n(Q_\phi) = 
{\mathbb E}
_{\mathcal D}
\left[
    \left( \sum_{k=0}^{n-1}\gamma^k r_{t+k} + \gamma^n \hat{Q}^\pi(s_{t+n},a_{t+n})- Q_{\phi}(s_t,a_t) \right)^2 \right]
\end{equation}

The total critic loss becomes:
\begin{equation}   
    \mathcal{L}_{\text{Critic}}(Q_\phi) = \mathcal{L}_1(Q_\phi) + \lambda_n \mathcal{L}_n(Q_\phi)
\label{loss_critic}
\end{equation}

\bigbreak
\textbf{Behaviour-Cloning loss.}

Following SAC-BC \cite{nair2018overcoming}, we use a behaviour-cloning loss to train the actor. This loss prevents the policy from deviating too much from the demonstrations, and accounts for sub-optimimality of the demonstrations by filtering out updates where the critic under-performs.

\begin{equation}
\notag
\label{sacbc}
\mathcal{L}_\text{BC}(\pi_\theta) = \sum_{i \in \text{Demo}} ||\pi_\theta(s_i)-a_i||_2^2 \mathds{1}_{Q_\phi(s_i,a_i)>Q_\phi(s_i,\pi_\theta(s_i))}
\end{equation}

The total actor loss becomes:
\begin{equation}
    \mathcal{L}_{\text{Actor}}(\pi_\theta) = \mathcal{L}(\pi_\theta) + \lambda_\text{BC} \mathcal{L}_\text{BC}(\pi_\theta)
\label{loss_actor}
\end{equation}

\bigbreak
\textbf{Modified reward derived from Advantage Learning.}

Following SAIL \cite{{DBLP:journals/corr/abs-2012-11989}}, we use a modified reward that promotes self-imitation. The idea is to increase the current action-value for actions whose returns are unexpectedly good. The new reward can also be smaller than the original reward, compensating also for negative experiences. Let $G_t = \sum_{t'=t}^T \gamma^{t'-t}r_{t'}$ be the return-to-go at each time-step. The modified reward is:

\begin{equation}
\label{reward_sail}
\Tilde{r}
(s_t,a_t) = r(s_t,a_t) + \lambda_\text{SAIL} \left[ \max \left( G_t, Q_{\phi'}(s_t,a_t) \right) - \hat{V}(s_t) \right] 
\end{equation}

For DDPG, since the policy is deterministic, we can directly use $\hat{V}(s_t) = Q_{\phi'}(s_t, \pi_{\theta'}(s_t))$ as the value-function estimator. For SAC, we use $\hat{V}(s_t) = V_{\psi'}(s_t)$ where $V_\psi$ is a separate network. Since the first version of the SAC algorithm had a value-function network, we choose the same loss function to train $V_\psi$:

\begin{equation}
\notag
    \mathcal{L}(V_\psi) = {\mathbb E}_{s \sim {\mathcal D}}\left[\left(V_{\psi}(s) - 
    {\mathbb E}_{a \sim \pi_\theta(\cdot|s)} \left[ Q_\phi(s,a) - \alpha \log \pi_\theta(a|s)  \right]
    \right)^2\right]
\label{sac-v}
\end{equation}

The new reward $\Tilde{r}$ replaces the standard reward in the Bellman equation for the critic update. For instance, for DDPG the critic loss becomes:

\begin{equation}  
\notag
\mathcal{L}(Q_\phi) = \underset{(s_t,a_t,r_t,s_{t+1}) \sim {\mathcal D}}{{\mathbb E}}\left[
    \left( \Tilde{r}_t + \gamma \hat{Q}^\pi(s_{t+1},a_{t+1})- Q_{\phi}(s_t,a_t) \right)^2 \right]
\end{equation}

\RestyleAlgo{ruled}
\begin{algorithm}
\SetAlgoLined
 \caption{STIR$^2$}
Require: $L$ amount of steps to relabel.

Set $b$ according to equation \ref{condition}.

 Initialize buffer with demonstrations, set reward $r=b$ for the last $L$ non-final transitions.
 
 Initialize empty episode.
 
 \While{not converged}{
  sample a batch $\mathcal{B}$ from the buffer
  
  \uIf {$b > 0$}{
  $b \leftarrow \text{decay\_rule}(b)$
  }\Else{
  set $r = 0$ for all non-final transitions in $\mathcal{B}$
  }
  
  do off-policy RL update (one step of gradient descent following equations \ref{loss_critic}, \ref{loss_actor} and \ref{reward_sail})

  \uIf{len(\text{episode}) == 0}{
   collect one episode
   
     \uIf{episode is successful}{
     
   
   set $r = b$ for the last $L$ non-final transitions
   }
   }
   pop a transition $(s,a,s',r)$ from the episode and add it to the buffer
 }
\end{algorithm}


\section{Experimental setup}
\label{experimental_setup}


We test our method on two different simulation environments: RLBench \cite{james2020rlbench} and Meta-World \cite{yu2019meta}. For both environments, an episode ends once the robot has completed the task, or after expiration (from 50 to 100 time-steps depending on the task). 
The reward is fully sparse, and is equal to $+100$ if the robot solves the task and 0 otherwise. 

\bigbreak
\textbf{RLBench}.
We evaluate our method on four simulated tasks for a 6-degrees-of-freedom robot manipulator: reaching a ball, pushing a button, flipping a switch, and sliding a block to a target square.
The target object (ball, button, switch, block and square) appears randomly at the beginning of each episode within the reach of the robot: the ball appears anywhere in the 3D space, the button, block and square are bound to the tabletop, and the switch is on a wall which appears randomly in the scene facing the robot.
The state has $19+x$ dimensions, 19 from the robot proprioceptive state (joint angles, joint speeds, gripper pose) and $x$ from the task-related information (3D coordinates of each target object). 
For the more contact-based sliding task, the robot receives 3D end-effector position commands (fixed orientation). For the other tasks, the robot receives 6-dimensional joint speed commands.

\bigbreak
\textbf{Meta-World}.
We evaluate our method on four simulated tasks for a 7-degrees-of-freedom robot manipulator: reaching a ball, pressing a button, closing a drawer, and sliding a block to a target location.
The target object (ball, button, drawer, block) appears randomly at the beginning of each episode within the reach of the robot: the ball appears anywhere in the 3D space, while the drawer, (horizontal) button, and  block are bound to the tabletop.
We refer to the Meta-World paper for the action and state spaces which we do not change.

\bigbreak
\textbf{Demonstrations}.
For RLBench, the expert demonstrations are provided by the environment and rely on OMPL \cite{sucan2012open} for motion planning. For Meta-World, we first train an RL agent for each task from dense rewards, and retain a batch of its successful episodes at test-time as demonstrations.

\bigbreak
\textbf{Experiments}.
The basic SAC algorithm without demonstrations is able to solve these tasks on some runs. Our goal is to solve the tasks consistently and reduce the amount of training steps required to do so.
We want to answer three questions: 
How does our method perform? 
How robust is it to the hyper-parameters?
How does it perform under rougher conditions (weaker base algorithm, few demonstrations available, no demonstrations available)?

In order to answer to these questions, we compare our methods SAC-R$^2$ and SAC-STIR$^2$ to a simple baseline SAC-Demo, which we define as SAC with demonstrations in the buffer, as well as SAC-fD \cite{DBLP:journals/corr/VecerikHSWPPHRL17}, SAC-BC \cite{nair2018overcoming} and SAC-SAIL \cite{DBLP:journals/corr/abs-2012-11989}. Unless stated otherwise, we set $L=10$, $\lambda_\text{SAIL} = 0.9$, $\lambda_\text{BC} = 2$, $\lambda_n = 1$ and $n = 5$. 

For SAC-R$^2$, we use the decay presented in equation \ref{pseudo-decay}. If the success rate doesn't improve over a period of 5000 iterations, a secondary linear decay is applied to completely erase the bonus. We always choose the largest bonus allowed by the condition \ref{condition}.

We apply the following modifications to all 6 algorithms:
\begin{itemize}
\item For all methods except SAC-BC: Single buffer initially filled with 200 demonstrations, and kept thereafter at a ratio of $10\%$ demonstration data. We decide not to set a limit on the number of unique demonstrations available, meaning that around 500-1000 more demonstrations are added to the buffer (depending on the task and number of iterations). For SAC-BC: Two separate buffers, one exclusively filled with demonstrations (with a comparable amount).
\item We additionally put 1000 random interactions alongside the demonstrations in the buffer, and pre-train during 3000 iterations before collecting any data.
\item For all methods except SAC-fD: The data is sampled from the buffer according to prioritized experience replay (PER) \cite{DBLP:journals/corr/SchaulQAS15}. For SAC-fD: A modified version of PER which boosts the sampling of demonstration data is used.
\item The replay ratio on the collected data is set to 32. Since the batch size is set to 64, the agent takes two environment steps per training step.
\item L2 regularization losses on the weights of the critic and the actor.
\end{itemize}

On top of these changes, SAC-BC uses an auxiliary behaviour-cloning loss and resets some episodes to demonstration states, SAC-fD uses an auxiliary n-step loss, SAC-R$^2$ uses the reward bonus presented in Section \ref{method}, and SAC-SAIL uses a modified reward (see (\ref{reward_sail})).

All the results are smoothed with a rolling window of 100 episodes, and the standard error is computed on three random seeds.

\pagebreak


\section{Experimental Results}
\label{results}

\begin{figure}[h!]
\centering
   \begin{subfigure}[b]{0.49\textwidth}
    \includegraphics[width=\textwidth]{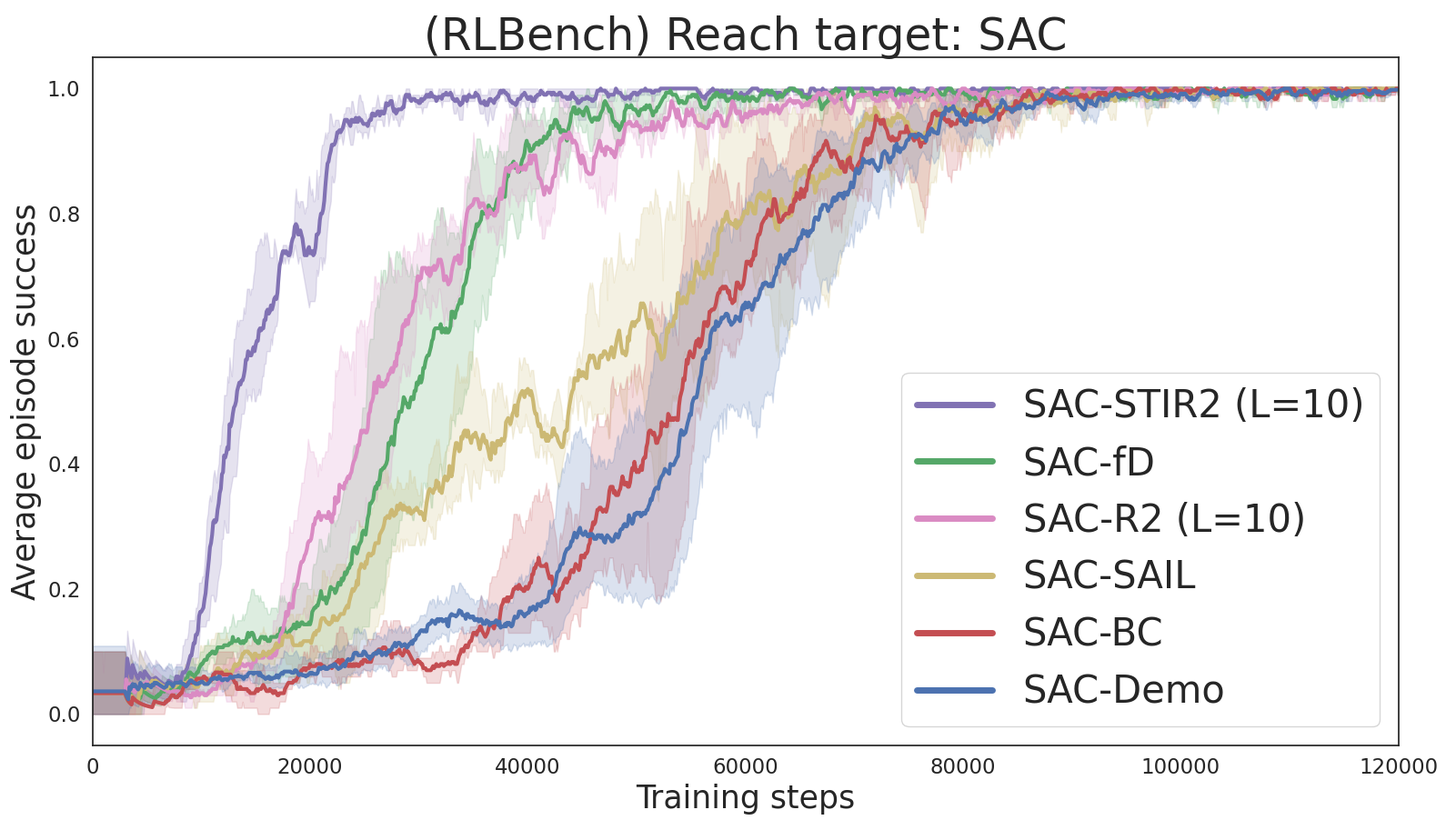}
    \caption{} \label{reach:sac}
  \end{subfigure}
  \begin{subfigure}[b]{0.49\textwidth}
    \includegraphics[width=\textwidth]{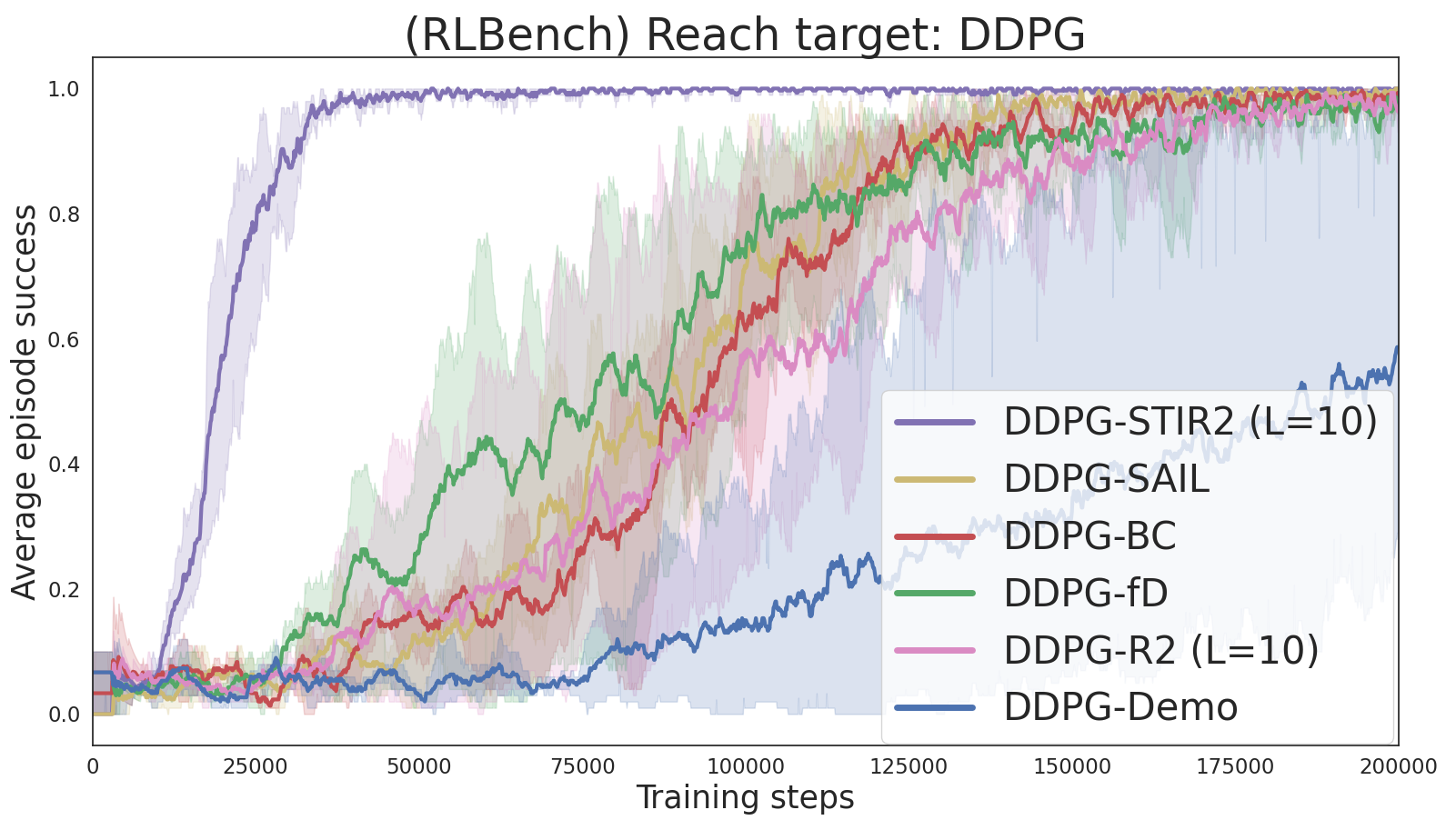}
    \caption{} \label{reach:ddpg}
  \end{subfigure}
  \caption{Learning curves for the RLBench reaching task.}
  \label{reach}
\end{figure}

\begin{figure}[h!]
\centering
  \begin{subfigure}[b]{0.49\textwidth}
    \includegraphics[width=\textwidth]{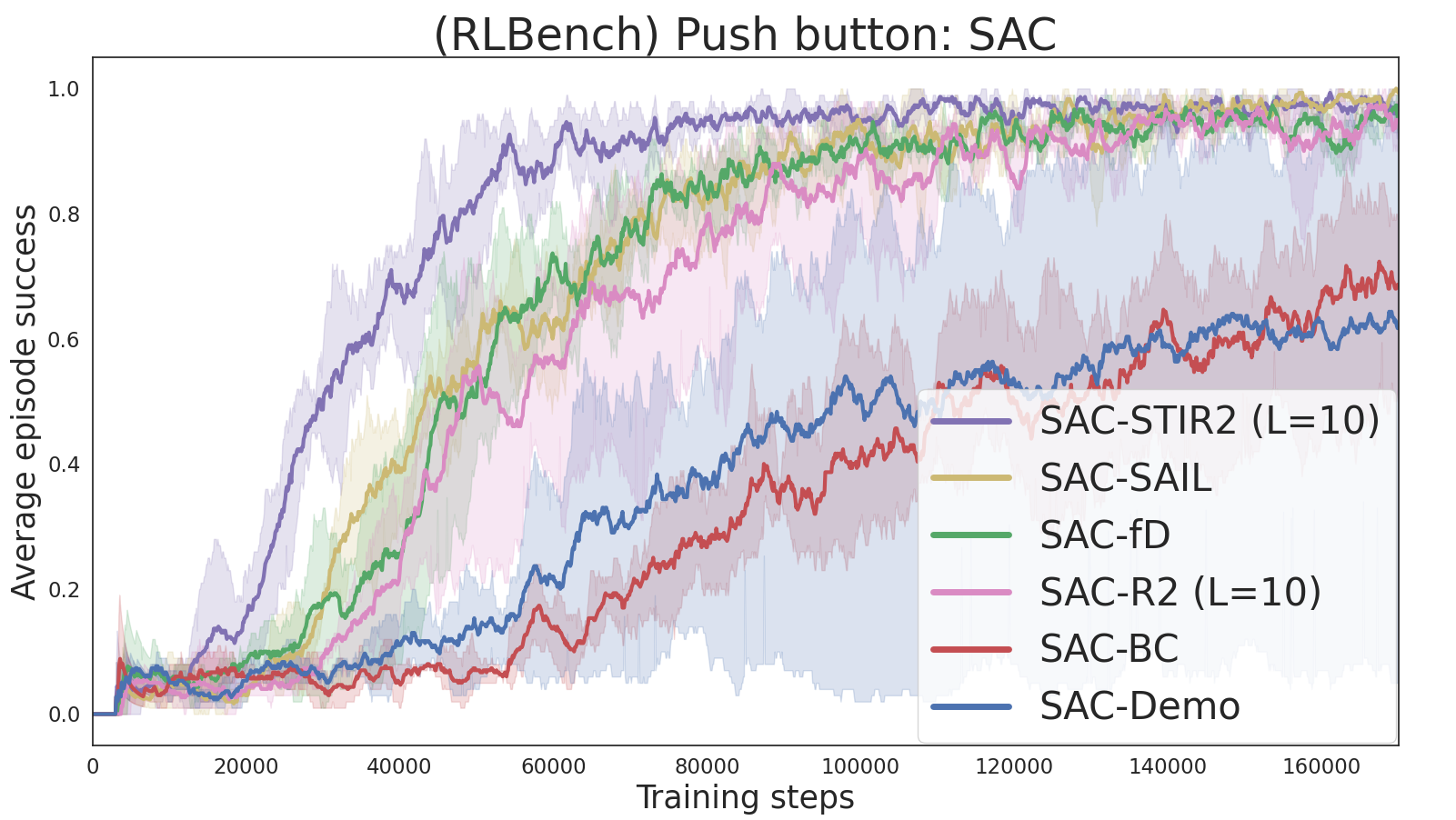}
    \caption{} \label{curves:push}
  \end{subfigure}
  \begin{subfigure}[b]{0.49\textwidth}
    \includegraphics[width=\textwidth]{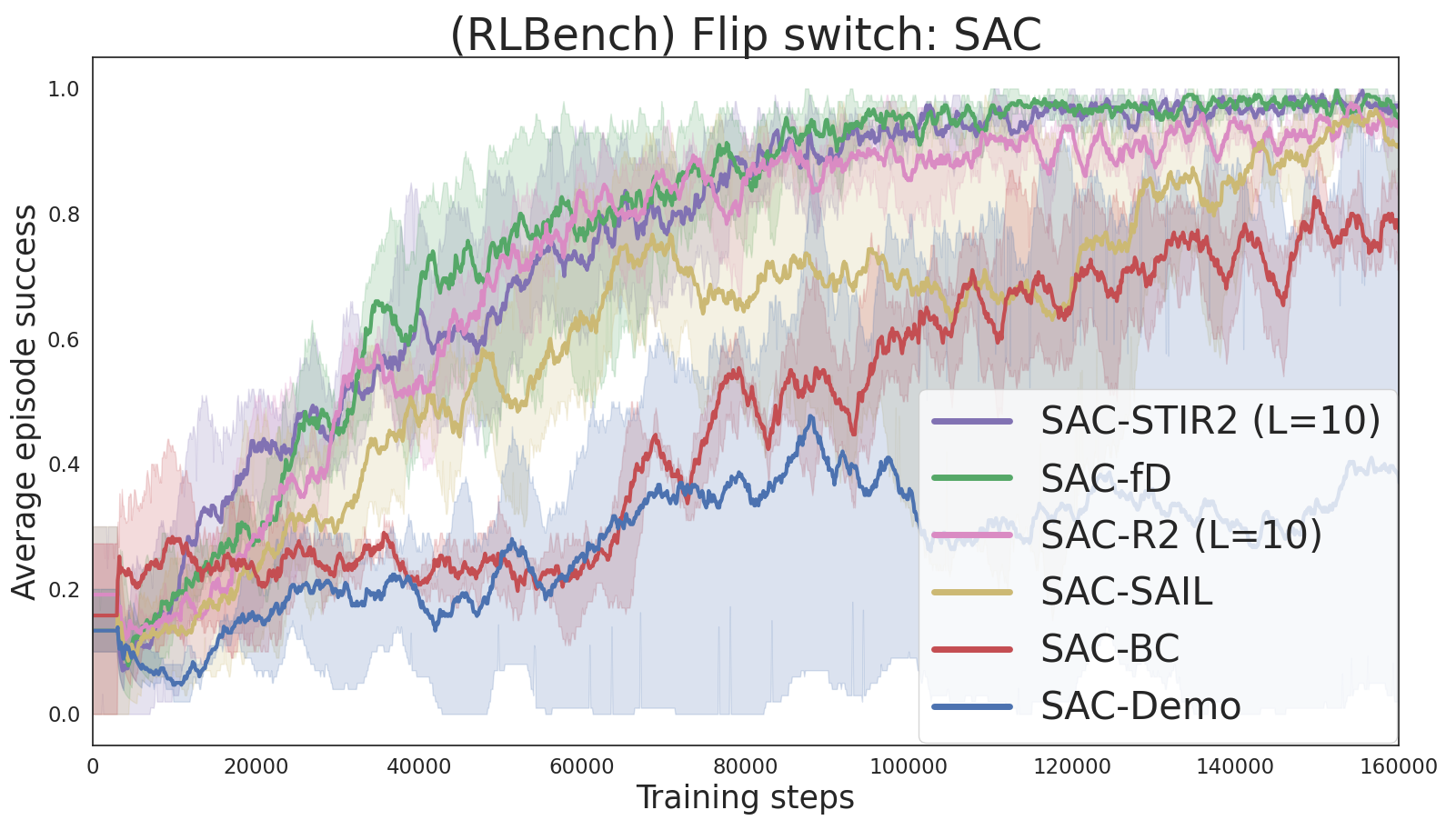}
    \caption{} \label{curves:press}
  \end{subfigure}
  \begin{subfigure}[b]{0.49\textwidth}
    \includegraphics[width=\textwidth]{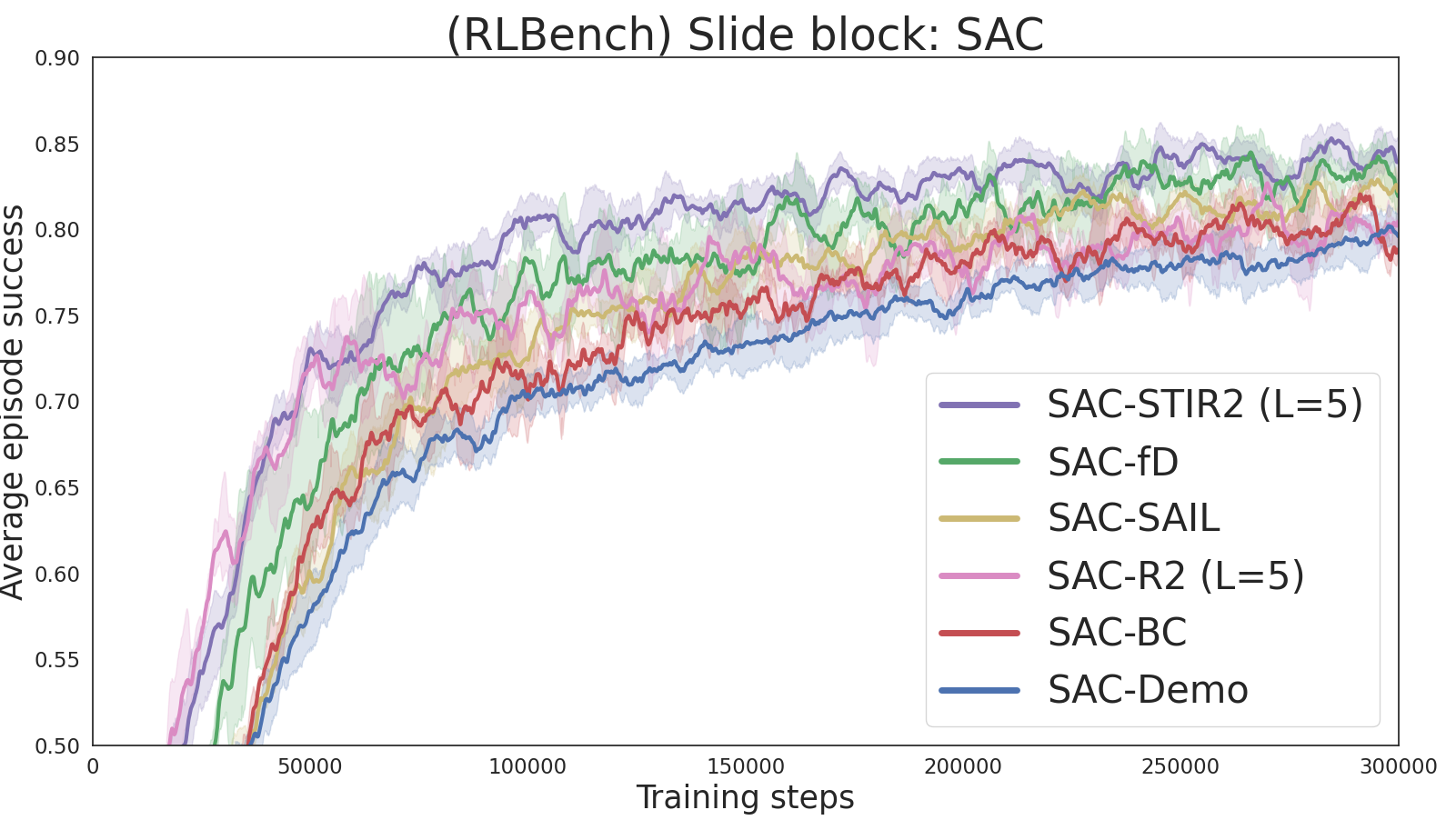}
    \caption{} \label{curves:slide}
  \end{subfigure}
  \caption{Learning curves for three RLBench tasks.}
  \label{curves}
\end{figure}

\begin{figure}[ht]
\centering
  \begin{subfigure}[b]{0.49\textwidth}
    \includegraphics[width=\textwidth]{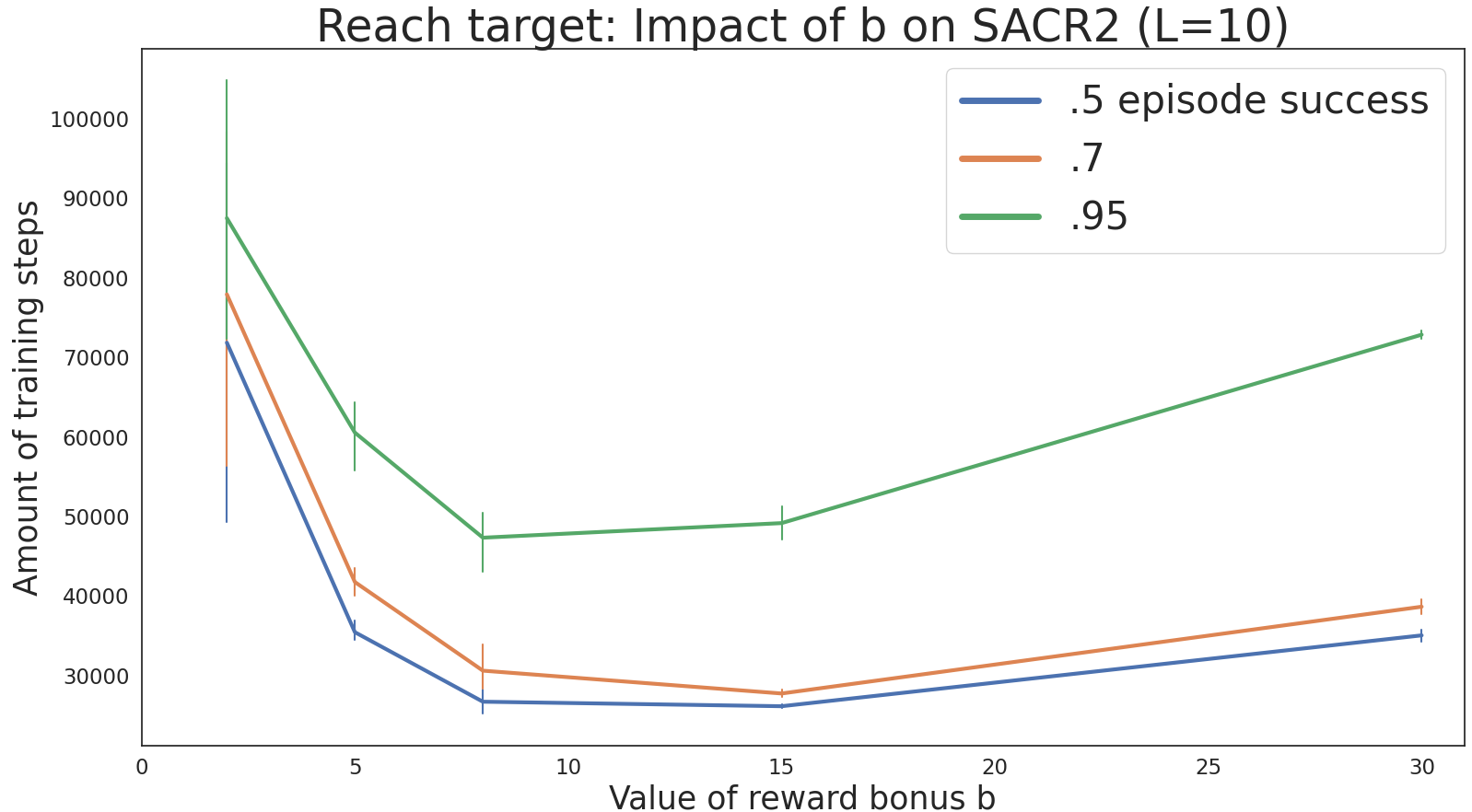}
  \end{subfigure}
  \begin{subfigure}[b]{0.49\textwidth}
    \includegraphics[width=\textwidth]{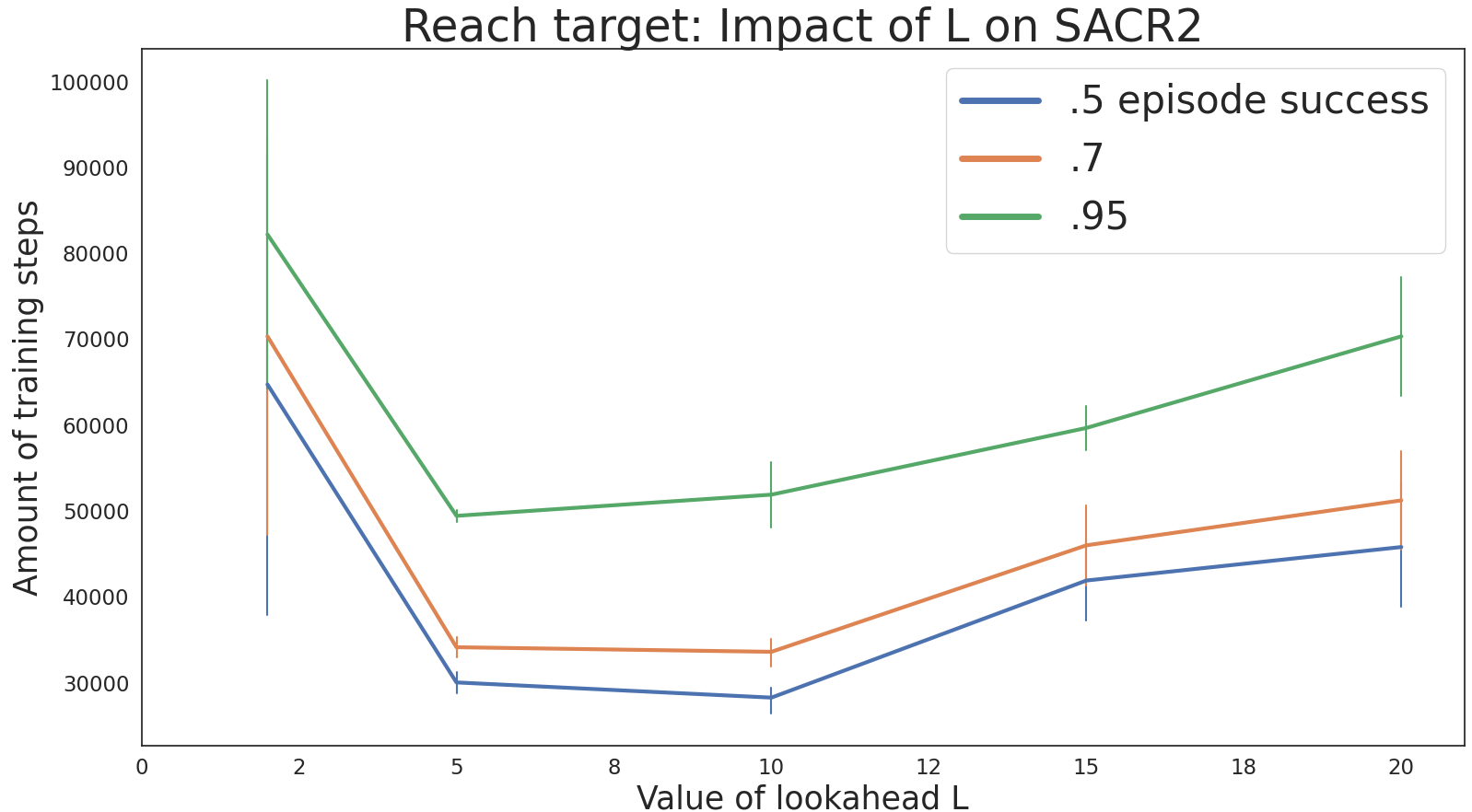}
  \end{subfigure}
  \caption{For a given hyper-parameter value, we plot the number of training steps required for the average episode success to reach a certain threshold.
}
  \label{params}
\end{figure}

\subsection{RLBench Experiments}

The results (Figures \ref{reach:sac}, \ref{curves:push}, \ref{curves:press}, \ref{curves:slide}) show that SAC-STIR$^2$ is significantly better than the other methods on the \textit{reach target} and \textit{push button} tasks: It is 1.6 times faster to reach a relative success rate of $90\%$. It also has the best performance on the \textit{flip switch} task, but is tied with both SAC-R$^2$ and SAC-fD. On the more difficult \textit{slide block} task, SAC-STIR$^2$ is able to reach a higher success rate than the other methods ($\sim 84\%$, compared to $\sim 82\%$ for second-best SAC-fD and $\sim 78\%$ for SAC-Demo).
Figure \ref{reach:ddpg} shows that DDPG-R$^2$, DDPG-BC, DDPG-fD and DDPG-SAIL have similar performances in the \textit{reach target} task, while DDPG-Demo struggles and DDPG-STIR$^2$ outpaces everybody by a factor of 4 to reach a relative success rate of $90\%$.


\bigbreak
\textbf{Impact of hyper-parameter L}.
Figure \ref{params} (bottom) shows the impact of different values of $L$. Having a longer window hurts the performance, as adding a bonus to transitions closer to the goal is probably more important. Also, if $L$ is bigger than the length of an episode some bonuses will not be assigned. For this task, there seems to exist an optimal value somewhere between 5 and 10.

\bigbreak
\textbf{Impact of hyper-parameter b}.
What happens if we ignore the condition found in Section \ref{theory} and choose $b$ as an independent hyper-parameter? 
Figure \ref{params} (top) shows the impact of different values of $b$ for a fixed $L=10$.
Surprisingly, violating the condition ($b>8$) does not hinder the performance that much, which means that as long as there is some sort of decay the value of $b$ is not that important. The drop in performance only becomes noticeable with very large values. 
As expected, choosing a smaller bonus ($b<8$) reduces the performance as the algorithm becomes closer to SAC-Demo (which can be seen as SAC-R$^2$ with $b=0$).

\begin{figure}[h!]
\centering
  \begin{subfigure}[b]{0.49\textwidth}
    \includegraphics[width=\textwidth]{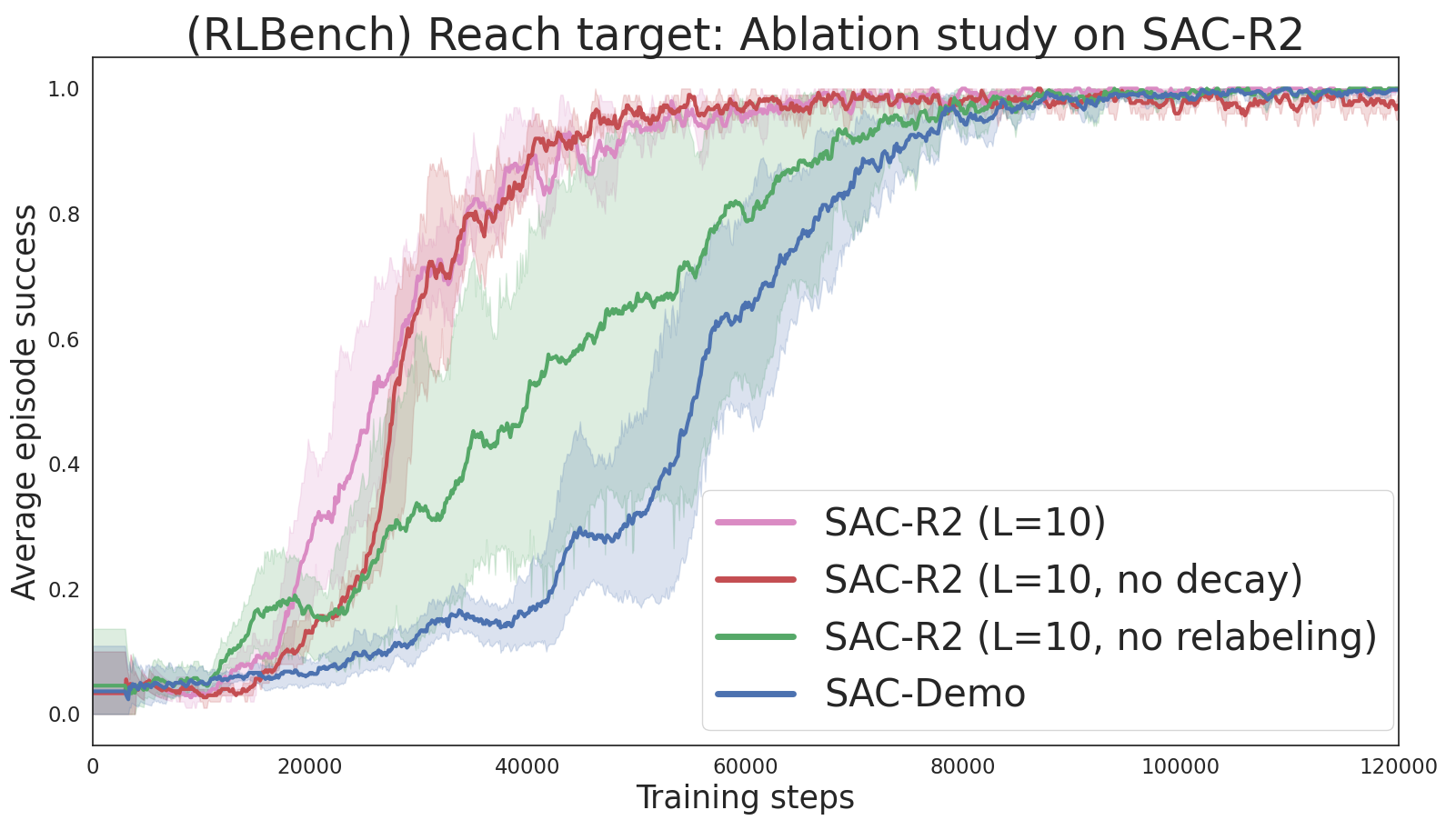}
    \caption{} \label{ablation:a}
  \end{subfigure}
  \begin{subfigure}[b]{0.49\textwidth}
    \includegraphics[width=\textwidth]{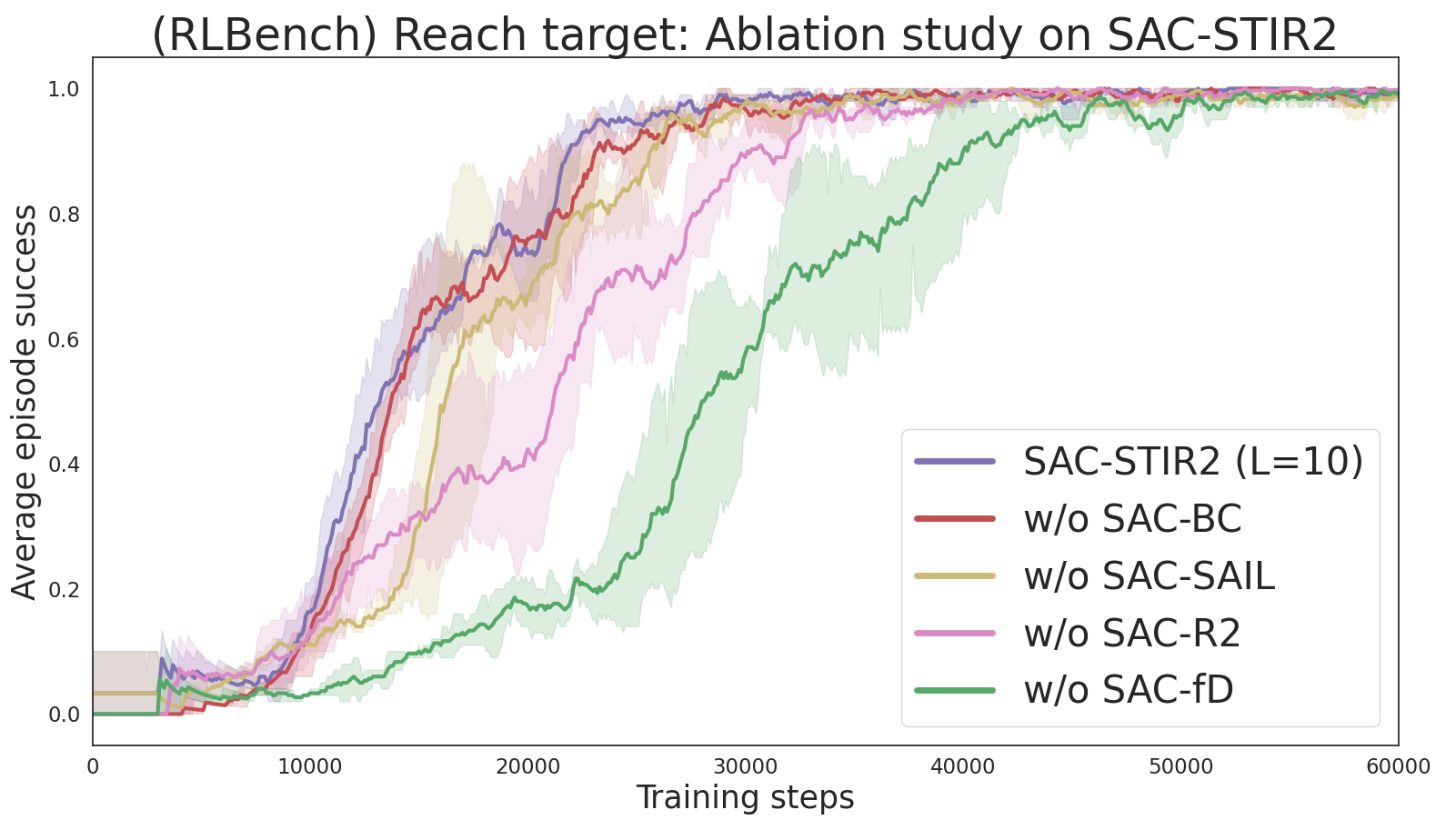}
    \caption{} \label{ablation:b}
  \end{subfigure}
  \caption{Learning curves for different ablation studies.}
  \label{ablation}
\end{figure}

\begin{figure}[h!]
\centering
  \begin{subfigure}[b]{0.49\textwidth}
    \includegraphics[width=\textwidth]{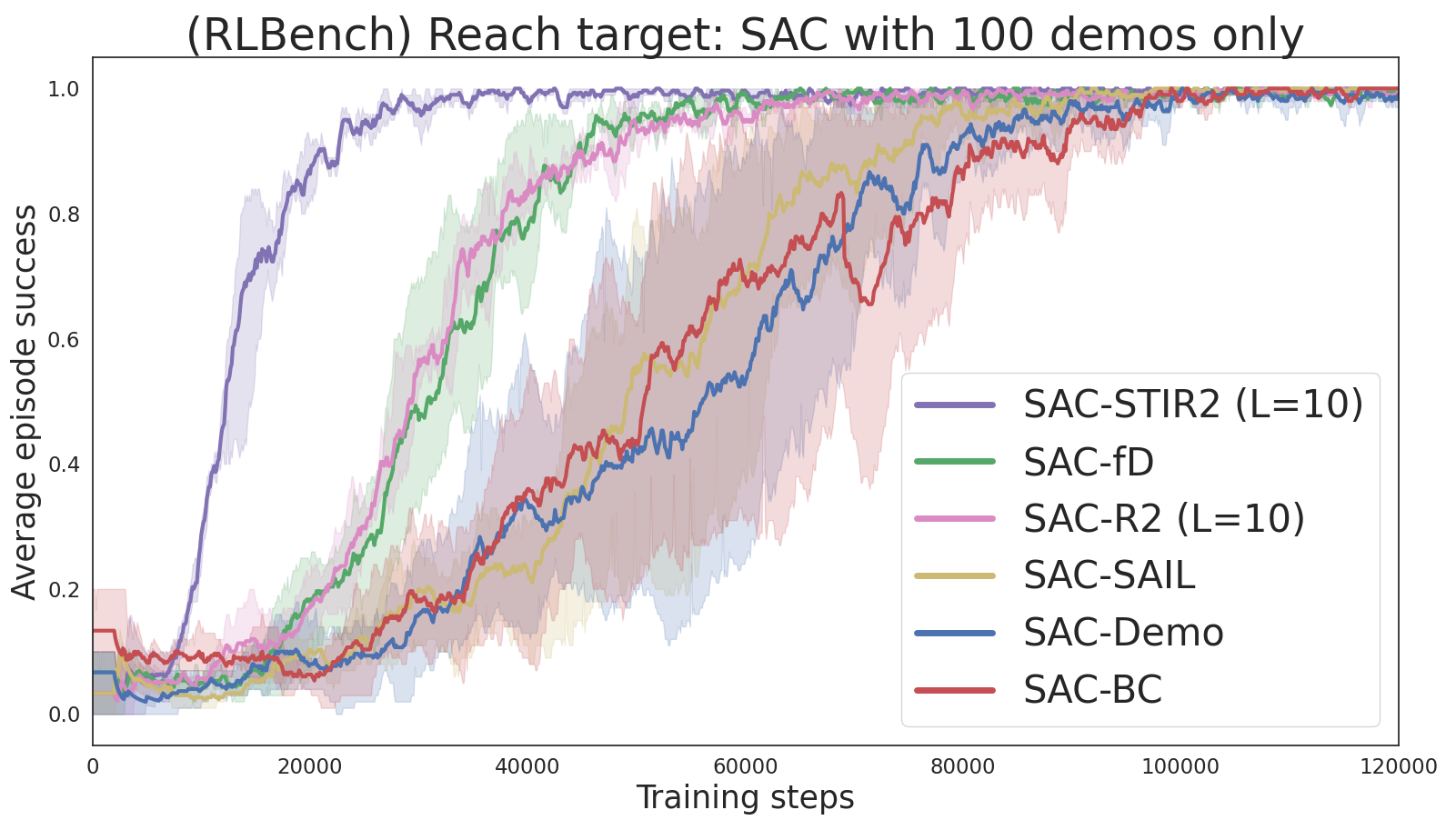}
    \caption{} \label{demos:100}
  \end{subfigure}
  \begin{subfigure}[b]{0.49\textwidth}
    \includegraphics[width=\textwidth]{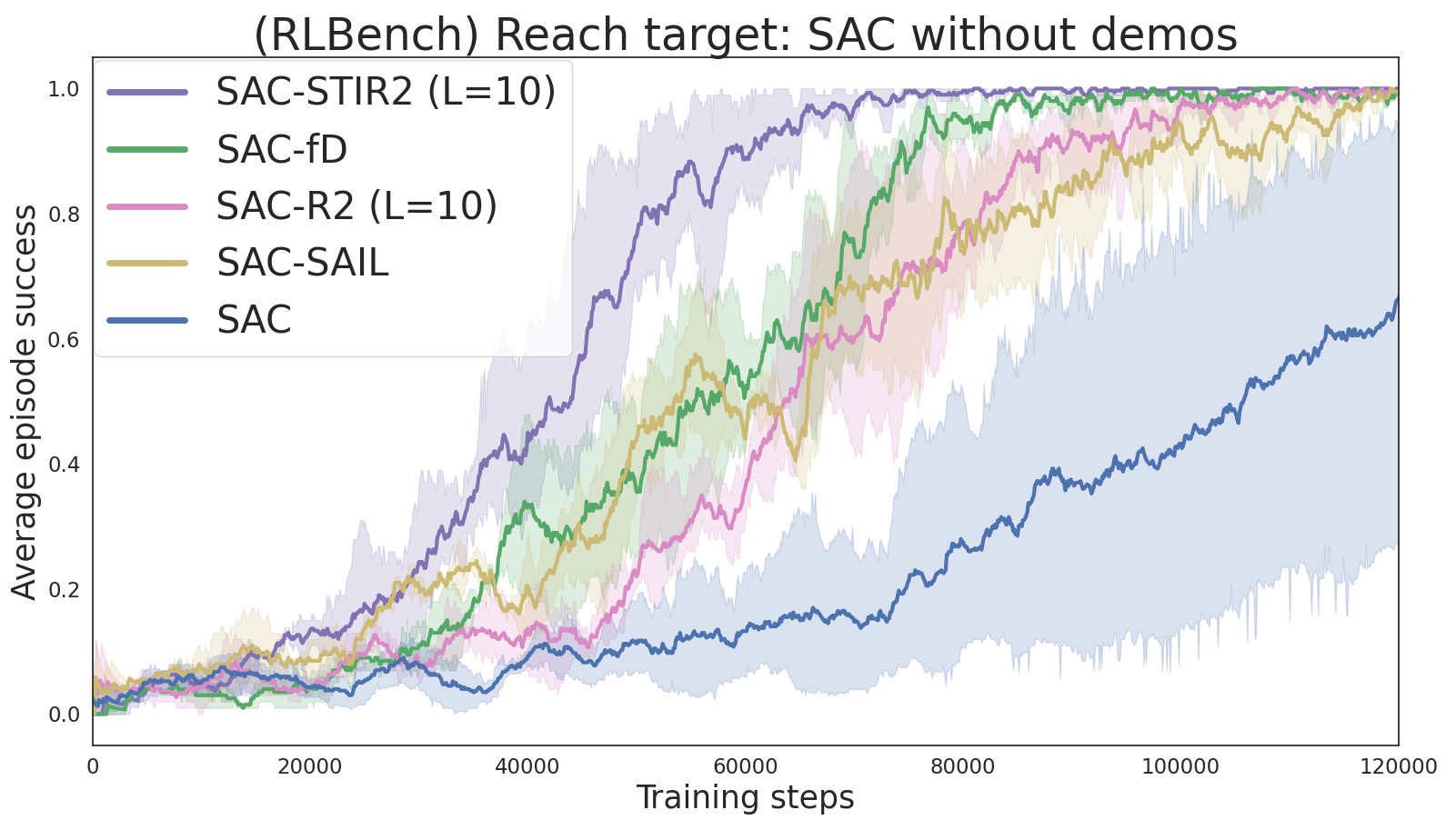}
    \caption{} \label{demos:0}
  \end{subfigure}
  \caption{Learning curves for the RLBench reaching task (low-data regime).}
  \label{demos}
\end{figure}

\bigbreak
\textbf{Ablation study on SAC-R$^2$}.
Figure \ref{ablation:a} shows the impact of online relabelling and decay.
What happens if we skip the former? We still give a reward bonus to the transitions coming from demonstrations, but we no longer relabel successful episodes. The green curve shows
that, without relabeling, the learning process seems more unstable, probably due to the lack of consistency in the rewards once the agent has collected enough successful episodes. 
The red curve shows the impact of not decreasing the bonus during training. Both curves (pink and red, with and without decay) are very similar, but the agent without decay does not actually reach a $100\%$ success rate, and additional results show that the larger the bonus, the smaller the final performance. As discussed in Section \ref{theory}, SAC-R$^2$ does not seem to converge to the optimal policy without a decay.

\bigbreak
\textbf{Ablation study on SAC-STIR$^2$}.
Figure \ref{ablation:b} shows that the contributions of all four methods (SAC-R$^2$, SAC-BC, SAC-fD and SAC-SAIL) are important to SAC-STIR$^2$, since removing either of them causes the performance to drop.
For this task, SAC-fD is clearly the most important element, followed by SAC-R$^2$. SAC-BC performs particularly bad on this task (Figure \ref{reach:sac}), so it is unsurprising that it has the lesser impact.




\bigbreak
\textbf{Low-data regime}.
The previous results were obtained with a large amount of demonstrations, which is unrealistic for real-world applications. What happens if we limit the amount of available demonstrations to just 100? Figure \ref{demos:100} shows that the results are very similar (but slightly worse for all methods) to the high-data regime (Figure \ref{reach:sac}).

\bigbreak
\textbf{Learning without demonstrations}. 
Finally, can SAC-R$^2$ also improve the performance when no demonstrations are available, by just relabeling successful episodes? Figure \ref{demos:0} shows similar results to those with demonstrations in the buffer: SAC-STIR$^2$ performs the best, while the weakest SAC baseline is only able to solve the task on some runs. Note that SAC-BC is not included in the figure because its behaviour-cloning loss relies on demonstration data.

\subsection{Meta-World Experiments}

\begin{figure}[h!]
\centering
  \begin{subfigure}[b]{0.49\textwidth}
    \includegraphics[width=\textwidth]{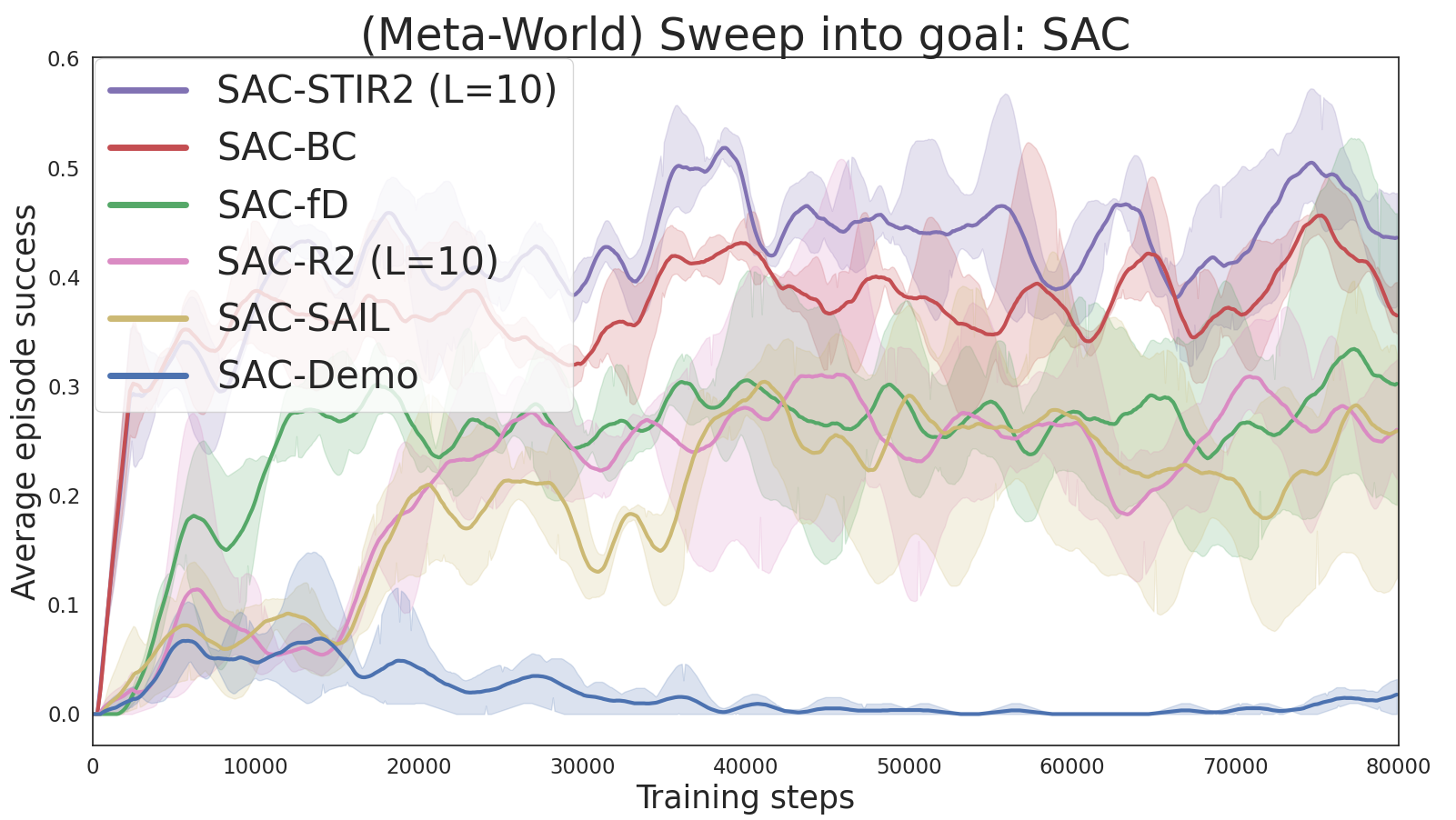}
  \end{subfigure}
  \begin{subfigure}[b]{0.49\textwidth}
    \includegraphics[width=\textwidth]{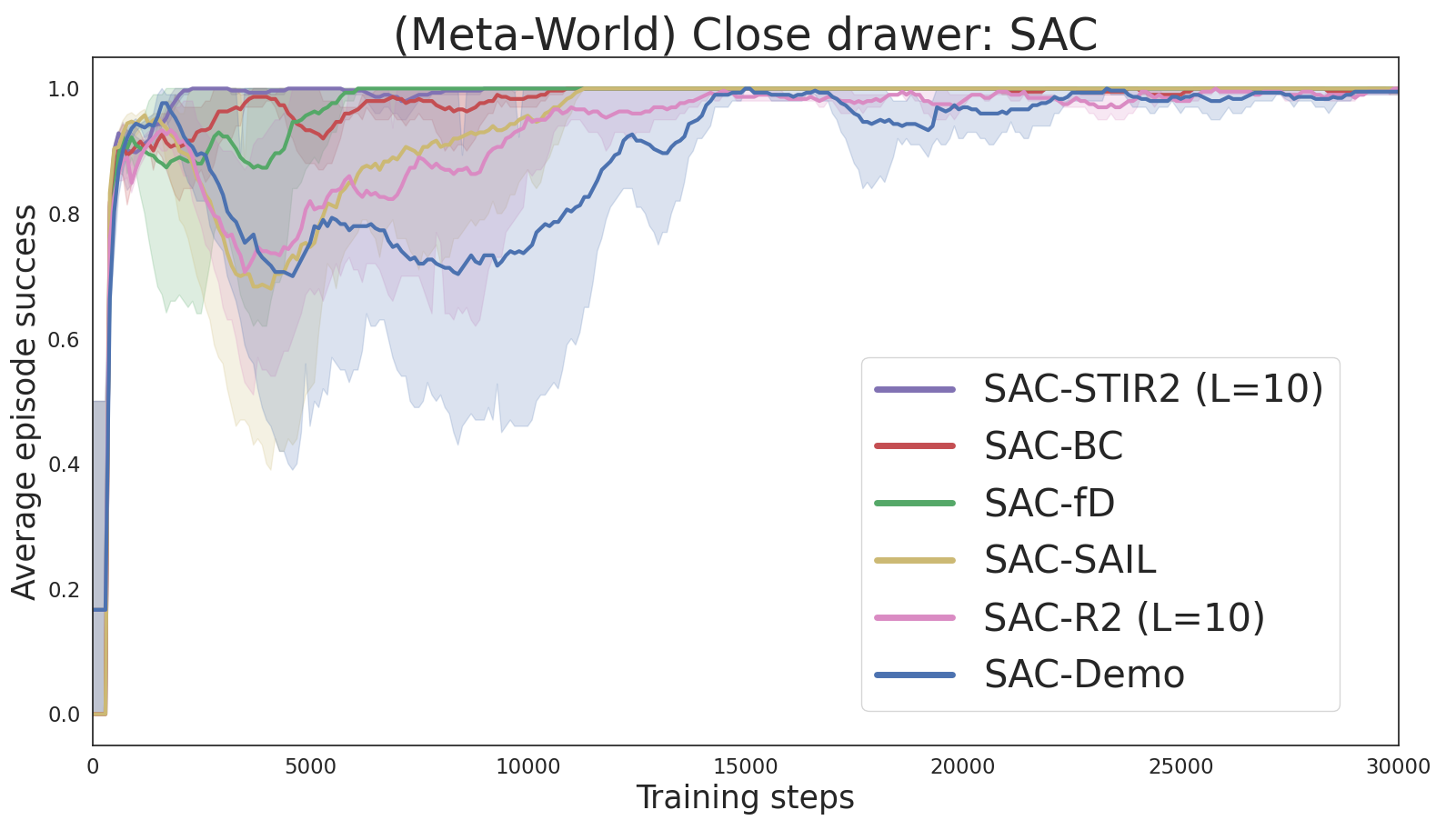}
  \end{subfigure}
  \begin{subfigure}[b]{0.49\textwidth}
    \includegraphics[width=\textwidth]{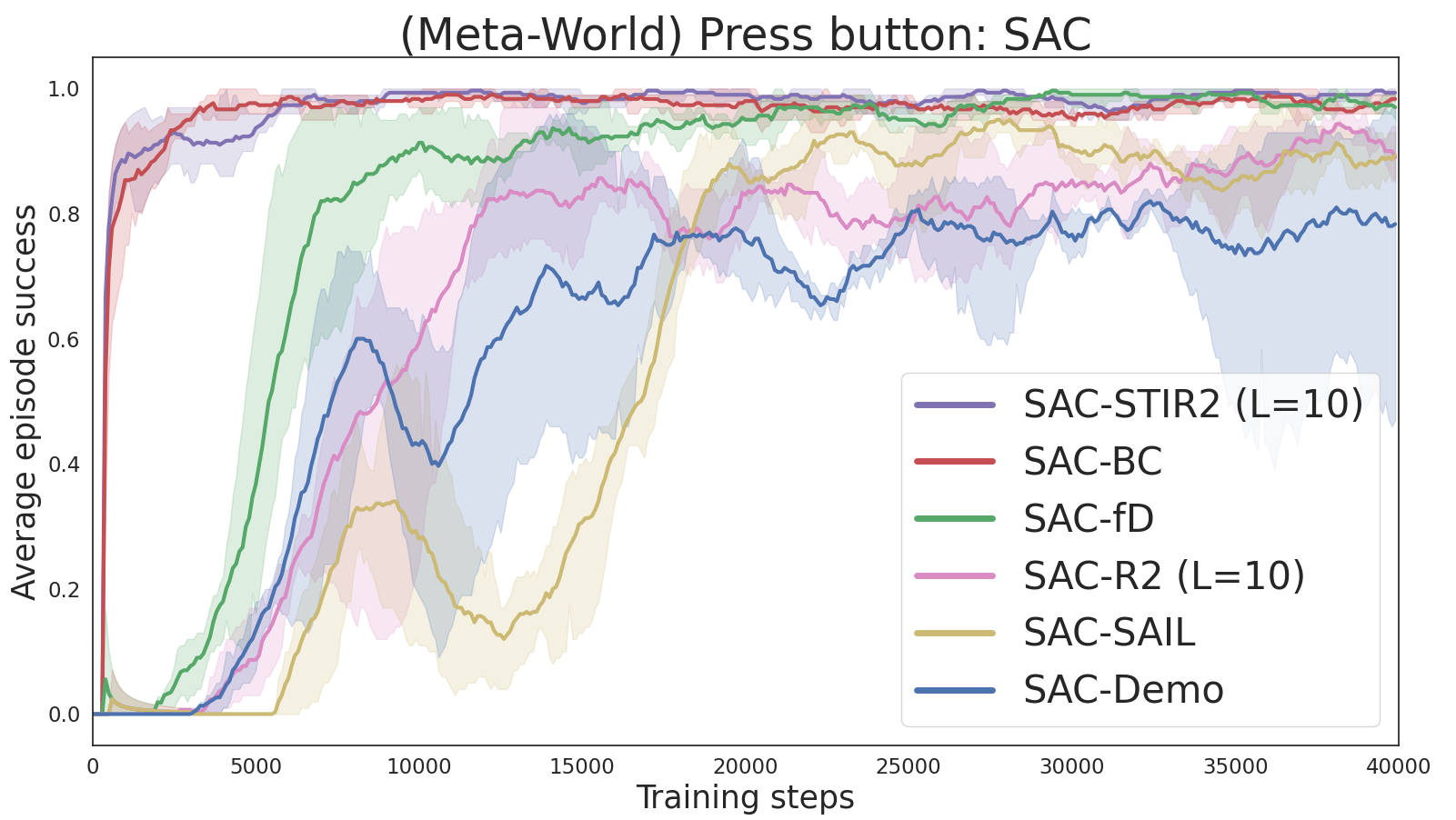}
  \end{subfigure}
   \begin{subfigure}[b]{0.49\textwidth}
    \includegraphics[width=\textwidth]{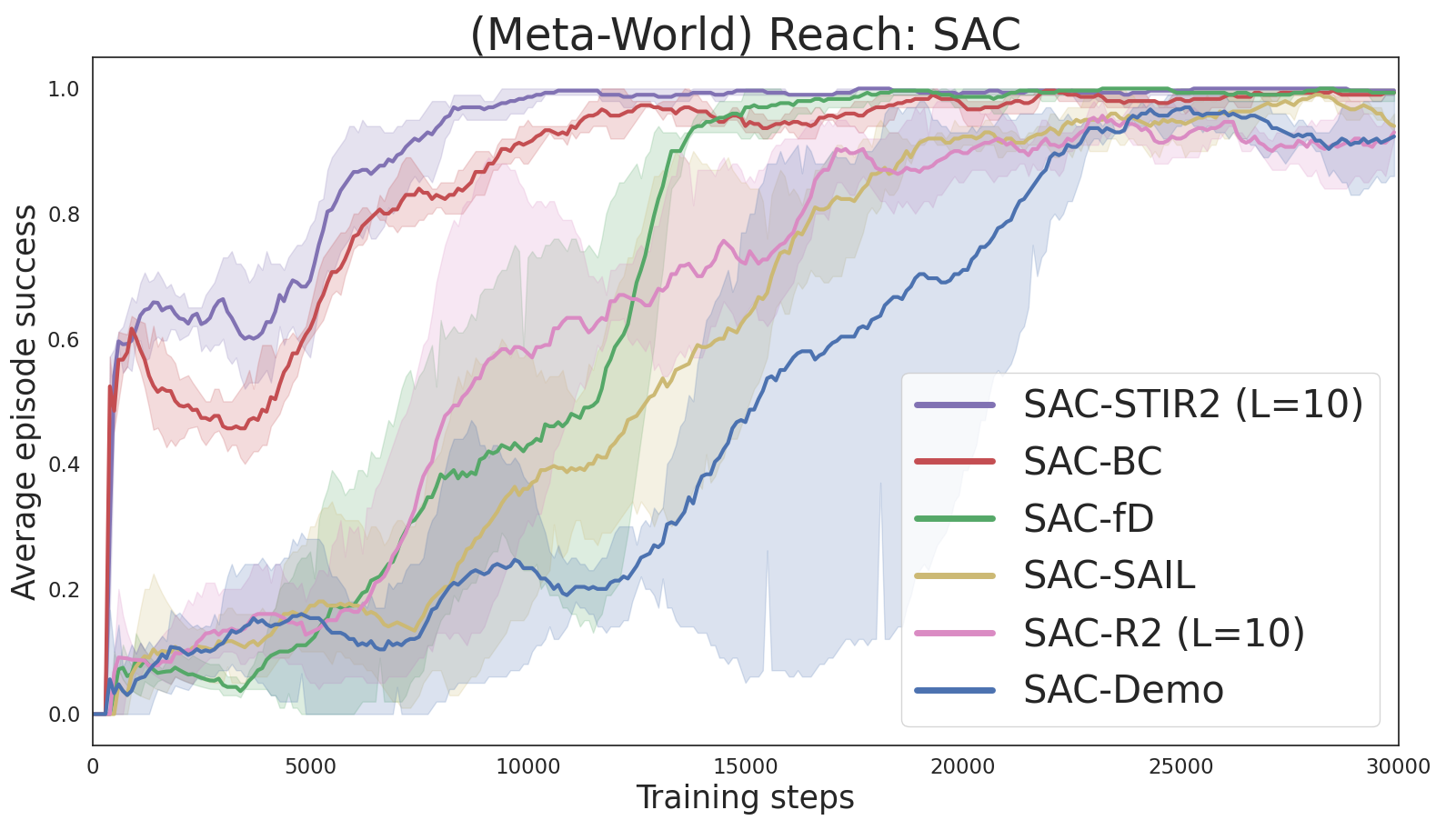}
  \end{subfigure}
  \caption{Learning curves for four Meta-World tasks.}
  \label{metaworld}
\end{figure}

The results (Figure \ref{metaworld}) show that SAC-STIR$^2$ is also the best method across the four Meta-World tasks. It is faster to converge on the easier tasks, and attains a higher final performance (almost one point higher) on the more difficult \textit{sweep into goal} task. However, the overall results are only marginally better than those of SAC-BC, which is the strongest among all baselines. Surprisingly, SAC-BC was one of the weakest baselines on the RLBench tasks. We believe this is tied to the origin of the demonstrations. It is more straight-forward for the agent to imitate the actions of an expert which is also an RL agent, while the motion-planning demonstrations from RLBench might exhibit more un-natural behaviours.



\section{Discussion}

We propose Reward Relabeling (R$^2$), a generic method that can be applied to any off-policy RL algorithm in any sparse-reward environment. It encourages two behaviours: imitate the expert demonstrations (if available), and imitate the past successful trajectories.
From our experiments, our method SAC-R$^2$ had comparable results to SAC-fD \cite{DBLP:journals/corr/VecerikHSWPPHRL17}, SAC-SAIL \cite{DBLP:journals/corr/abs-2012-11989}, and SAC-BC \cite{nair2018overcoming}. 
We also propose Self and Teacher Imitation by Reward Relabeling (STIR$^2$), a new algorithm which combines our reward-relabeling method with the n-step loss from SAC-fD, the behaviour-cloning loss from SAC-BC, and the modified reward from SAIL. We show that these methods stack together, as the best results were obtained with SAC-STIR$^2$ on all tasks. Similar results were obtained with DDPG as the base algorithm, and in theory it could be implemented on top of any continuous-action off-policy RL algorithm.


As ongoing and future work, we are currently evaluating our algorithms on more tasks. We eventually want to test our method on more complex tasks from both benchmarks, such as sequential tasks or tasks where designing a dense reward would be more challenging. We also want to compare against more baselines, in particular a pure imitation-learning algorithm such as Implicit BC \cite{florence2022implicit}.
Finally, it would be interesting to test our method with a Q-learning-type base algorithm on a task with discrete actions. 


\pagebreak

\newpage
\phantom{dsdfsf}
\newpage






\bibliographystyle{ACM-Reference-Format} 
\bibliography{sample}


\end{document}